\documentclass{llncs}

\usepackage{makeidx}

\title{Socially Impaired Robots: Human Social Disorders and Robots' Socio-Emotional Intelligence}

\author{
    Jonathan Vitale
    \and
    Mary-Anne Williams
    \and
    Benjamin Johnston
}

\institute{
  Innovation and Enterprise Research Lab, QCIS\\
  University of Technology Sydney,
  Ultimo, NSW 2007, Australia\\
  \email{Jonathan.Vitale@student.uts.edu.au}\\
  \email{Mary-Anne@TheMagicLab.org}\\
  \email{Benjamin.Johnston@uts.edu.au}\\
 }
 
\begin{document}

\maketitle

\begin{abstract}
  
  Social robots need intelligence in order to safely coexist and interact with humans. Robots without functional abilities in understanding others and unable to empathise might be a societal risk and they may lead to a society of socially impaired robots. In this work we provide a survey of three relevant human social disorders, namely autism, psychopathy and schizophrenia, as a means to gain a better understanding of social robots' future capability requirements. We provide evidence supporting the idea that social robots will require a combination of emotional intelligence and social intelligence, namely  \emph{socio-emotional intelligence}. We argue that a robot with a simple socio-emotional process requires a simulation-driven model of intelligence. Finally, we provide some critical guidelines for designing future socio-emotional robots.
  
\end{abstract}

\section{Introduction}
\label{sec:introduction}

Social robots are embodied intelligent agents designed to coexist and interact with humans or with other social robots \cite{dautenhahn1999bringing}. In order to avoid risks for the society their behaviour must to be safe and conform to social norms. Social intelligence, defined as the ability to make sense of others' actions and react appropriately to them \cite{langton2000eyes}, plays a crucial role in regulating acceptable interactions between people. Thus, social robots will require a form of social intelligence too \cite{williams2012robot}.

In psychology studies, together with social intelligence, we find a subtly different form of intelligence, namely emotional intelligence. \emph{Emotional intelligence} is defined as the ability to perceive, manage, and reason about emotions, within oneself and others \cite{ermer2012emotional}. 
Emotional intelligence is generally considered part of social intelligence \cite{salovey1989emotional}. However, in the robotic literature social intelligence is commonly introduced without considering this important need of emotional states elicitation and understanding.

In our everyday society there are people with an abnormal social behaviour. This population exhibits brain disorders involving deficits in social intelligence. It seems plausible that future social robots with high level cognitive capabilities, but lacking in social intelligence skills will develop similar social deficits \cite{williams2012robot}. In this work we want to provide a survey of human social disorder concerning deficiencies in social intelligence, so to gather significant information that can be used to trace a design guideline for future social robots. This is necessary in order to avoid the possibility of developing a society of socially impaired robots.


How can people achieve social intelligence? If we turn to psychology, philosophy, or the cognitive sciences in general, Theory of Mind (ToM) is the most shared and common strategy for gaining social intelligence abilities. ToM is defined as the ability to attribute mental states to oneself and others \cite{premack1978does}. It provides mechanisms for comprehending/explaining everyday social situations, for predicting and anticipating others' behaviour, and even for manipulating other individuals. 

Two main models of ToM are provided in the literature: Theory-Theory (TT) and Simulation Theory (ST).
In the TT account the mind-reader deploys a na\"ive psychological theory to infer mental states in others from their behaviour, the environment, and/or their other mental states \cite{goldman2005simulationist}. On the other hand, according to ST, the mind-reader select a mental state to attribute to others after reproducing or enacting within himself the very state in question, or a relevantly similar one \cite{goldman2005simulationist}. In this way the mind-readers do not need theories; instead, they use their own body as a model of others.



Both ST and TT provide valid theories of how people can master the ability of making sense of others' social behaviour. However, from a social perspective there is a remarkable difference: ST requires an embodiment and the use of phenomenological mechanisms in order to ``put the mind-reader in other's shoes'', however, TT does not. ST contributes in \emph{resonating} and so \emph{empathising} with others, thus modulating socially acceptable behaviours, whereas TT works in a more mechanical and `cold' vision. More specifically, a simulation-driven approach provides phenomenological bases for the development of \emph{social reputation}, since it allows us to think about what others think of us and feel the corresponding positive or negative sentiment (i.e. they see me as a \emph{good} person vs. \emph{bad} person). The feeling elicited from the social reputation process via simulation, in turn, might become an incentive for individuals to conform to social norms \cite{izuma2011insensitivity}.

Individuals with disorders related to social deficits show difficulties in empathising with and mentalising about others \cite{farrow2007empathy}. As suggested by Dautenhahn: ``the better we understand human psychology and human internal dynamics, the more we can hope to explain embodiment and empathic understanding on a scientific basis'' \cite[p. 22]{dautenhahn1997could}. Following this advice, and motivated by the previously exposed problematic scenario, in Section \ref{sec:survey} we propose a brief survey of the most documented human social disorders and related deficits functional to social intelligence, namely autism, psychopathy and schizophrenia. In Section \ref{sec:discussion} we discuss this evidence, we define \emph{socio-emotional intelligence} and then relate it to necessary sub processes. Finally, in Section \ref{sec:conclusions} we provide our conclusions and a research agenda that will lead to the development of socio-emotional robots. To our knowledge this is the first study that tries to investigate human social disorders as a means to provide design principles for future social robot development.

\section{A Survey of Human Social Disorders}
\label{sec:survey}

\subsection{Autism Spectrum Disorders}

Autism Spectrum Disorders (ASD) individuals are characterized with deficits in three main areas: (i) communication, (ii) social interaction, and (iii) restrictive and repetitive behaviours and interests \cite{american2000diagnostic}. 

ToM is one of the main social skills in which ASD population shows deficiencies \cite{farrow2007empathy}. Compared with normally developed persons, ASD individuals are poorer at reasoning about what others think, know, or believe, recognizing emotional expressions and gestures, and making social attributions and judgements \cite{bachevalier2006orbitofrontal}. However, deficits in ToM ability are only the tip of the iceberg. 

Indeed, ASD individuals show other social deficits functional to ToM, for example they are poor in understanding the emotional content of face expression, gestures and vocalizations and they fail in using these social signals as a way to express their own internal emotional state (e.g. arm around shoulder, hand over mouth, signalling embarrassment, \dots) \cite{brothers2002social}.
These deficits in emotion recognition/responding often lead to an impoverished facial affect. Thus, ASD individuals are perceived as unable to feel emotions. However, studies with electrodermal responses and self-report measures suggest that ASD individuals have appropriate emotional responsiveness to others \cite{dziobek2008dissociation}. Hence, ASD individuals seem to be able to normally experience such phenomenological internal states at least.

One of the earliest signs of ASD is a lack of sensitivity to social cues. For example, they exhibit poor eye contact, they have difficulties in joint attention (either using eyes, head pose or pointing gestures) and they show disinterest to other people \cite{typical2012cognitive,brothers2002social,farrow2007empathy}. Many studies investigated the gaze direction of ASD individuals using eye tracking systems. They found that this population look less at the eyes relatively to control participants (for a review see \cite{sasson2006development}). Birmingham and her research group suggest that perhaps the abnormality in ASD people lies in the likelihood that they will seek out and select social information from a scene; if such a population does not consider important social signals like others' gaze orientation, they will not be able for example to infer where others are looking. 
Due to this perceptual deficit, they might have less evidence to use during a mind-reading process \cite{typical2012cognitive}.

Aligned to this perspective, Dawson \emph{et al.} propose that social orienting deficits might cause ASD development and subsequent ToM deficiencies \cite{dawson1998children}. This hypothesis suggests that individuals developing with ASD fail to attend social stimuli from an early age. This lack of crucial social information during the normal development provokes later social cognitive deficits, such as facial expression processing and mind-reading.

Indeed, social orienting deficits can also explain their lacks in emotion recognition from social signals. 
Perhaps this population possesses embodied mechanisms to feel others' emotions, as well as mechanism for `resonating' to others facial expressions and body movements (i.e. a functioning Mirror Neuron Systems, see \cite{bons2013motor,iacoboni2006mirror} for a discussion). However, they lack of a social reward process and they cannot direct the attention on stimuli necessary for promoting mind-reading abilities \cite{izuma2011insensitivity}. Without focusing on important social signals, like the eyes, they might have severe limitations in ToM \cite{bons2013motor}. 

\subsection{Psychopathy}

The World Health Organization classifies \emph{Psychopathy} as a form of antisocial (or dissocial) personality disorder \cite{world1992icd}. Characteristics of such disorder are: (i) callous unconcern for the feelings of others; (ii) incapacity to maintain enduring relationships, though having no difficulty in establishing them; (iii) very low tolerance to frustration and a low threshold for discharge of aggression, including violence; (iv) incapacity to experience guilt or to profit from experience, particularly punishment; (v) marked proneness to blame others, or to offer plausible rationalizations, for the behaviour that has brought the patient into conflict with society \cite{world1992icd}.

In contradistinction to what is commonly believed, psychopathic individuals do not always present violent and criminal behaviour. Indeed, this population is mainly characterized by a lack of `emotional empathy' \cite{farrow2007empathy}; they have a reduced ability to feel other people's emotional state, especially sadness and fear \cite{decety2007empathic}. Psychopathic subjects have deficits in moral emotions such as remorse and guilt and they are usually indifferent to shaming and embarrassing situations \cite{ermer2012emotional}. 

Antisocial personalities usually exhibit a poor executive control, that is normally necessary for socially appropriate conduct \cite{decety2007empathic}. This dysfunction might be due to non-responding violence inhibition mechanisms that are normally triggered during the feeling of others' distress in order to prevent the execution of antisocial behaviours \cite{blair1995cognitive}. Indeed, psychopathic individuals own a poor behavioural control, leading often to impulsivity. Furthermore, a study on startle reflex modulation of visual attention demonstrates that psychopathic individuals, compared to normal population, present an abnormal valence pattern \cite{patrick1993emotion}. The authors suggest that even if psychopaths express different subjective judgements to positive vs. aversive visual stimuli, they may find such stimuli equally inviting from an attention controlling perspective. This may be due to a dysfunction in attention reflex reactions when perceiving unpleasant content \cite{patrick1993emotion}. These evidences well support the existence of emotion regulation deficiencies in such population \cite{ermer2012emotional}.

Deficits in emotion regulation seem to affect also face processing abilities. Psychopaths are impaired when processing fearful, sad and disgusted facial expressions, whereas it seems that they do not have impairments with happy facial expression, even if this should be due to the ease with which such expression is recognized \cite{farrow2007empathy}. Furthermore, this population has deficits in other emotional processing skills, such as failure to show normal response differentiation to emotional and neutral words, and abnormal reactions to emotional stimuli and events \cite{ermer2012emotional}.

Surprisingly, this mental disorder does not involve abnormal levels of intelligence \cite{ermer2012emotional}. In fact, in contrast to other disorders like autism, psychopathic individuals successfully complete ToM tasks and currently there is no evidence of impairments in `cognitive' (i.e. not emotional or empathic) ToM ability \cite{farrow2007empathy}. However, due to deficits in experiencing emotions, psychopath individuals cannot simulate them, and must rely exclusively on cognitive inputs in order to fulfil a mind-reading task \cite{decety2007empathic}.

\subsection{Schizophrenia}

Schizophrenia is a severe psychiatric disorder altering emotional, cognitive, and social functions \cite{parasuraman1998attentive}. In particular, significant impairment in social functioning is considered one diagnostic characteristic of schizophrenia \cite{american2000diagnostic}. Such impairment can have seriously impacts on social relationships \cite{kennedy2012social}. Schizophrenic subjects suffer also from delusions and hallucinations, however in this survey we will consider only their deficits primarily related with social intelligence abilities (for a discussion on ToM and correlations with these symptoms, see \cite{brune2005theory}).

Similarly to autistic and psychopathic individuals, schizophrenic individuals lack general abilities in ToM and empathy \cite{sparks2010social,farrow2007empathy}. In fact, some current models of schizophrenia suggest that this disorder can be understood as a deficit in representing others' mental states (i.e. cognitive ToM) and of `resonating' to others' emotional states (i.e. empathy) \cite{farrow2007empathy}. However, even in this case deficits in these abilities are just the tip of the iceberg; indeed different cognitive sub processes seem to be affected in schizophrenia leading to a differentiation of such deficiencies respect to psychopathy and ASD. For example, whereas schizophrenic individuals seem to be able at least to understand the intended meaning of sincere interpersonal exchanges (differently from ASD population), they show deficiencies in insincere interactions, such as in understanding sarcastic conversations, that indeed lie more on emotional features such as prosody and intonation \cite{sparks2010social}.

Schizophrenic individuals exhibit blunted feeling and they usually have inappropriate affective responses in social situations \cite{farrow2007empathy}. They show abnormalities of skin conductance response and they mostly respond with negative affect (e.g. depression) \cite{farrow2007empathy}. Furthermore, this population exhibits deficits in subjective experience of emotion \cite{myin2000schizophrenia}. Studies demonstrate that schizophrenic patients emotionally respond with fewer positive and negative facial expressions in response to emotional stimuli compared to normal population \cite{earnst1999emotional}. However, evidence from other studies support the idea that schizophrenic people can indeed feel emotional states, but they cannot sustain attention over the emotional stimuli and thus maintaining such emotional state during time.

Horan \emph{et al.} used affective pictures as emotional stimuli in order to record the Event Related Potentials (ERP) of schizophrenic subjects \cite{horan2010electrophysiological}. The results show that schizophrenic individuals experience comparable amounts of similar emotions with respect to normal populations during the initial ERP components, but not in Late Positive Potentials (LPP). The authors suggest that this population may have functioning emotional response mechanisms, but a disruption in a later component associated with sustained attention processing of the observed emotional stimuli \cite{horan2010electrophysiological}. This inability to maintain  the correct emotional response over time is correlated with deficits in executive control. In fact, without a sufficient elicitation of emotional processing over time it becomes critical to guide future behavioural choices \cite{horan2010electrophysiological}.

The previous studies demonstrate attention deficits in schizophrenic populations. Indeed, these individuals perform poorly on nearly all tests of sensory and cognitive vigilance and some studies also demonstrate deficits in selective attention \cite{parasuraman1998attentive}. It has also been shown that there are abnormalities in eye movements during the scanning of emotional facial expressions. Similarly to ASD people, schizophrenic individuals look less at the eye region of the face \cite{farrow2007empathy,typical2012cognitive}. Again, in a similar way as in ASD populations, schizophrenic patients show partial gaze avoidance specific to human faces, whereas they do not avoid gaze when they look to non-human faces \cite{williams1974analysis}. However, a study by Sasson \emph{et al.} provides evidence for a differentiation of emotional processing deficits in schizophrenics and ASD people \cite{sasson2007orienting}. Their results demonstrate that, whereas autism and schizophrenia share an impairment in fixating social stimuli (i.e. avoidance of eye region), the schizophrenic individuals show a delay in orienting the gaze to informative emotional stimuli (in this study faces). Thus it seems that, whereas autistic populations fail in the specificity of selective attention concerning emotional stimuli, schizophrenic population exhibited a generalized orienting delay \cite{sasson2007orienting}.

Given this evidence it seems that schizophrenic individuals' inability to maintain sustained attention and their delay of selective attention over emotional stimuli are strictly correlated with deficiencies in social intelligence.

\section{A Socio-Emotional Robotic Intelligence}
\label{sec:discussion}

In the introduction we suggested the need for social robots to possess social intelligence. We identified ToM as a crucial strategy to achieve such intelligence. However, we also mentioned that we need to prevent a society of `socially impaired robots'. Under this perspective creating robots able to perfectly understand humans' intentions and react appropriately to them in a pure rational way is not enough. In fact, we have seen that psychopaths, that indeed are able to understand others' intentions,  are a risk for the society as they can use ToM to manipulate or hurt people because unable to empathize or to feel sense of guilt.

We provided evidence demonstrating that the elicitation and regulation of emotions (i.e. emotional intelligence) are crucial skills needed to avoid, for example, psychopathic traits. Thus, given the importance of exhibiting both social and emotional intelligences, we provide a clearer and explicit definition of socio-emotional robot. We define a \emph{socio-emotional robot}\footnote{In robotic literature the term socio-emotional robot was already widely used; however, to our knowledge, nobody provided an explicit definition of it.} as a robot able to direct attention over others' social behaviour, to make sense of it and to elicit correct \emph{emotional processes} regulating and learning the expression of its behaviours in order to conform to society's culture, ethics, morality and common-sense.

Making sense of others' social behaviour can be achieved using ToM, whereas eliciting emotional processes for behaviour regulation and development is more related with empathy. Indeed, empathy may be a central feature of emotionally intelligent behaviour and it can be used to relate positively to others, thus increasing life satisfaction, reducing stress and motivating altruistic behaviour \cite{salovey1989emotional}. More specifically, a person can feel what the other person is feeling and so behave conformably to past experiences related with very similar feelings. Furthermore, a socio-emotional robot might learn correct behaviour through direct experience of its emotional states. Thus, having processes for emotion elicitation might facilitate the learning of society norms through a first-person experience.

\subsection{The Need for a Simulative Mechanism}

At the beginning of this paper we mentioned two possible approaches to master ToM: simulation-driven approaches and theory-driven approaches. Although both the approaches can be an acceptable explanation for mind-reading ability, when we look at empathy (and more in general at emotional intelligence) simulation-driven approaches play a crucial role in allowing this ability, as for example Goldman contends \cite{goldman1992defense}. Further support comes also from neuroscience studies of Gallese \cite{gallese2005embodied}. Thus, it seems plausible that in order to exhibit emotional intelligence a simulative mechanism is needed. As our previous survey on human social disorders demonstrate the need of both social and emotional intelligence (i.e. socio-emotional intelligence) in order to avoid social disorders, we suggest the need of simulative mechanism in social robots. With this recommendation we are not saying that theories about the world are not necessary for a fully understanding of others, but rather that at some preliminary levels a simulation process is needed in developing empathy and promoting socio-emotional intelligence.


A simplified socio-emotional process can be described as: (a) detecting a social behaviour, (b) enacting a simulation process given such stimulus and allowing an \emph{as-if} internal representation of it, (c) activate an appropriate viscero-emotional internal state (again through a simulation process), (d) use past experience and theories in order to give an interpretation to the perceived stimulus, (e) properly regulate the appraisal of the emotional state and the expression of an appropriate behaviour through (c), (d) and other theories about culture, ethics, morality and common-sense.

In most cases robotic studies on social intelligence make use of only two of such processes, namely (a) and (d). In fact, aligned with a pure information-based approach, researchers make use of datasets of social stimuli (face expressions, gestures, etc.) in order to create theories or models (d) to use for the interpretation of new social stimuli (a).
Given the previous survey we can argue that this approach potentially leads to possible social disorders and thus justify the need of a simulation mechanism in socio-emotional robots.

We have seen that ASD individuals have a deficit in directing attention over social stimuli (a), thus leading to deficit in representing them internally (b). On the contrary they do not show deficits in activating appropriate visceral states (c), if properly stimulated. Deficits in (a) throughout their life lead to learn poor social-life theories (d) in conjunction with their visceral states (c). This might explain their inability in fulfilling a successful socio-emotional process, since they poorly direct attention over social stimuli, thus reducing evidence for a mind-reading process, and as they develop a poor learning of (d) given (c).

On the other hand, psychopathic subjects show normal capacities in perceiving and processing social stimuli (a,b), but they cannot elicit viscero-emotional states (c). This again might explain their ability in mind-reading people (d) using social evidences (a,b) but their deficits in regulating empathic and moral behaviours (e) because unable to empathise with others (c).

Finally, schizophrenic population suggests the importance of synchronizing the processes of socio-emotional intelligence. A delay or dysfunction in sustain and selective attention over social stimuli (a) might lead to non-synchronized or distorted internal and cognitive processes (b,c,d) over time. This in turn leads to deficiencies in mind-reading and emotion regulation (e).


\section{Conclusions and Guidelines for Socio-Emotional Robots}
\label{sec:conclusions}

In this work we motivated the need for robots to be able to coexist in human spaces avoiding risks and costs for the society. In order to understand better how to design such a kind of safe robots we proposed a brief survey of human social disorders. We proposed the need of emotional intelligence together with social intelligence, and in order to clarify better these necessary intelligences in robots we provided an explicit definition of socio-emotional robot. We suggested the need of socio-emotional intelligence in order to avoid socially impaired robots. We provided a simple model of a socio-emotional process and we used the evidences from the survey as a way to motivate the need of a simulation process in order to avoid social deficits.

Given the need of socio-emotional intelligence for development of future robots, we suggest an agenda of necessary future research. First, social roboticists will need to provide appropriate mechanisms of attention modulation over social stimuli (a). In order to fulfil this target we will need to understand the mechanisms underlying social rewarding of social stimuli. Second, some kinds of simulation models will be necessary in order to represent perceived stimuli and activate an appropriate internal response in the robot (b,c). The internal representation of the stimuli might require a mapping from the external multimodal representation to an internal unimodal one; this is necessary in order to integrate different modalities under a unique and more computationally tractable form of representation. This is similar to our capacity of mapping multimodal external stimuli to unimodal neural activations. Third, we will need learning mechanisms allowing the association between internal representation and appropriate mental attributions (d). A further process of decision making is then necessary in order to drive the robot's executive attention and regulate its behaviour (e).

We want to conclude mentioning some limitations of the current study. First, the proposed survey is limited to three social disorders. There are others syndromes and brain dysfunctions that worth a discussion and enrich our argument, but for the sake of simplicity we limited our work to the most investigated social dysfunctions. Second, studies on ASD individuals, psychopaths and schizophrenics are not so linear as proposed in this survey. There are many controversies and open questions, but in order to provide a readable manuscript and an argument easy to follow we proposed some limited studies about hypotheses commonly shared in the related literature. We are confident that studies like the one reported in this manuscript will allow a better understanding of the human brain. This in turn is an essential knowledge if we want to develop intelligent machines. 


\bibliographystyle{splncs03}
\bibliography{main}

\end{document}